\documentclass[10pt,a4paper]{article}

\usepackage[a4paper,margin=0.78in,top=0.70in,bottom=0.72in]{geometry}
\usepackage{newtxtext,newtxmath}
\usepackage[T1]{fontenc}
\usepackage{microtype}
\usepackage{xcolor}
\usepackage{graphicx}
\usepackage{booktabs}
\usepackage{tabularx}
\usepackage{array}
\usepackage{amsmath}
\usepackage{siunitx}
\usepackage{enumitem}
\usepackage{caption}
\usepackage{subcaption}
\usepackage{float}
\usepackage{fancyhdr}
\usepackage{titlesec}
\usepackage{tcolorbox}
\usepackage{tikz}
\usetikzlibrary{positioning}
\usepackage{hyperref}
\usepackage{url}
\usepackage{longtable}
\usepackage{multirow}
\usepackage{listings}
\usepackage{lastpage}
\usepackage{hyperref}
\usepackage[numbers]{natbib}

\definecolor{onebitteal}{RGB}{0,150,150}
\definecolor{onebitblue}{RGB}{54,82,112}
\definecolor{onebityellow}{RGB}{250,245,218}
\definecolor{onebitlight}{RGB}{238,248,248}
\definecolor{onebitgray}{RGB}{90,90,90}

\hypersetup{
  colorlinks=true,
  linkcolor=onebitblue,
  citecolor=onebitblue,
  urlcolor=onebitblue
}

\setlist[itemize]{leftmargin=1.4em,itemsep=2pt}
\setlist[enumerate]{leftmargin=1.6em,itemsep=2pt}

\titleformat{\section}
  {\Large\bfseries\color{onebitblue}}{\thesection}{0.6em}{}
  [\color{onebitteal}\titlerule]
\titleformat{\subsection}
  {\large\bfseries\color{onebitblue}}{\thesubsection}{0.6em}{}
\titleformat{\subsubsection}
  {\normalsize\bfseries\color{onebitblue}}{\thesubsubsection}{0.6em}{}

\titlespacing*{\section}{0pt}{14pt}{7pt}
\titlespacing*{\subsection}{0pt}{10pt}{4pt}

\newcommand{\WModel}{W1.58A16}
\newcommand{\Qwen}{Qwen3-4B}
\newcommand{\TWLA}{TWLA}
\newcommand{\EATQ}{E2M-ATQ}

\newcommand{\TernaryFootnote}{\footnote{Throughout this report, ``ternary'' is used as engineering shorthand rather than as a claim of a fully native single-plane ternary network. The evaluated model is a layer-wise mixed representation: most weights use one ternary plane, while a salience-selected subset uses a second ternary plane together with per-group continuous offsets/scales and GPTQ compensation. The measured effective weight-information budget is approximately 1.641 bits/weight; embeddings and the language-model head remain BF16.}}

\newtcolorbox{takeaway}{
  colback=onebitlight,colframe=onebitteal,boxrule=0.5pt,
  left=8pt,right=8pt,top=6pt,bottom=6pt,arc=1mm
}
\newtcolorbox{productbox}{
  colback=onebityellow,colframe=orange!65!black,boxrule=0.5pt,
  left=8pt,right=8pt,top=6pt,bottom=6pt,arc=1mm
}

\begin{document}

\begin{center}
{\fontsize{23}{27}\selectfont\bfseries\color{onebitblue}
Post-Training Ternarization of Qwen3-4B\\[2pt]
Capability, Effective Bit Budget, Storage Compression, and Deployment\par}
\vspace{5pt}
{\large\color{onebitteal}\bfseries CLOE V2.0 -- OneBit AI\par}
\vspace{5pt}
\begin{tcolorbox}[colback=onebitlight,colframe=onebitteal,boxrule=0.5pt,width=0.95\linewidth]
\centering
\textbf{\Qwen{} FP16}
$\;\longrightarrow\;$ \textbf{KOTMS}
$\;\longrightarrow\;$ \textbf{\EATQ}
$\;\longrightarrow\;$ \textbf{GPTQ compensation}
$\;\longrightarrow\;$ \textbf{\WModel}
\end{tcolorbox}
\vspace{3pt}
{\small A capability, representation, compression, and deployment study of a
4B-parameter pretrained language model under aggressive post-training low-bit conversion.}
\vspace{6pt}

Anirudh Malik \quad|\quad Poojith Devan \quad|\quad M Sparsh Mehra

\vspace{3pt}
{\small OneBit AI | Technical Research Report | August 2026}
\end{center}

\begin{takeaway}
\textbf{Executive takeaway.}
This study asks a practical question: how far can an already-trained \Qwen{} checkpoint be pushed toward an ultra-low-bit representation without full retraining, and what remains useful once representation, capability, storage, and execution are measured separately?

Using the published \TWLA{} pipeline components, with weight-only A16 execution, the resulting \WModel{}\TernaryFootnote{} retains substantial performance on several commonsense and context-use tasks while losing more on knowledge-heavy benchmarks. Across the ten scored comparisons in the final benchmark table, FP16 averages 64.5\% accuracy and \WModel{} 54.7\%, a mean cost of 9.8 percentage points. WikiText-2 perplexity increases from 13.639 to 18.748, while PTB and C4 expose a wider 1.30--1.46$\times$ degradation range.

The representation itself is substantially smaller than FP16 once the quantizer's lattice is persisted: the effective weight budget is 1.641 bits/weight and the later lossless packing record reports a 3.96~GiB artifact versus an 8.29~GiB pre-packed checkpoint. However, the current execution path does not establish a speed advantage. The central systems lesson is therefore explicit: \textbf{compression is demonstrated; deployment acceleration remains an open engineering problem.}
\end{takeaway}

\begin{abstract}
Ultra-low-bit language models are attractive because reducing weight precision can reduce persistent storage, memory bandwidth requirements, and the hardware resources required to move model parameters. Yet a nominal ``1.58-bit'' label does not by itself describe the representation actually stored, the amount of pretrained capability retained, or the runtime behavior of the resulting artifact.

We study an end-to-end post-training conversion of \Qwen{}, an instruction-tuned 4B-parameter model, using the KOTMS rotation, E2M-ATQ ternarization, and GPTQ-style error compensation components of TWLA. The experiment is deliberately weight-only: activations remain at 16-bit precision and ILA-AMP is therefore skipped. The analysis emphasizes effective bit accounting, task-level capability retention, cross-corpus perplexity, calibration sensitivity, checkpoint composition, and deployment measurements.

The final evaluated conversion uses 1.641 effective bits/weight for the quantized linear weights, with 81.62\% of model parameters targeted. On the ten scored capability comparisons, accuracy falls from 64.5\% to 54.7\%. The losses are not uniform: BoolQ retains 84.6\% chance-corrected teacher performance, while ARC-Challenge retains 43.8\%. The pattern is consistent with stronger degradation on knowledge-heavy tasks than on contextual plausibility tasks. Perplexity rises from 13.639 to 18.748 on WikiText-2, 24.700 to 31.992 on PTB, and 19.831 to 28.966 on C4.

The study also exposes an important deployment distinction. A later packing run persists the actual ternary planes and scales and reports 8.29~GiB to 3.96~GiB with essentially unchanged perplexity. A separate third-party packing attempt was lossy and is explicitly excluded from the primary artifact claim. The packed artifact has not been benchmarked end-to-end for task accuracy or generation throughput in this source set. A preliminary Triton GEMV microbenchmark is 4.6$\times$ slower than FP16 cuBLAS on one tested shape. We therefore do not claim that compression alone yields faster inference.
\end{abstract}


\newpage

\section{Introduction}

The deployment cost of a language model is not determined only by the number of parameters. Transformer architectures themselves establish the dense-matrix foundation on which these compression methods operate.\cite{vaswani} The numerical representation of those parameters affects checkpoint size, memory traffic, loading time, accelerator utilization, and the feasibility of local or edge deployment. This has made low-bit model representations an active area of research, from conventional 8-bit and 4-bit PTQ through binary and ternary architectures.\cite{gptq,awq,smoothquant,bitnet,bitnet158}

Ternary representations are particularly attractive because a single symbol drawn from $\{-1,0,+1\}$ has information content
\begin{equation}
b_T=\log_2(3)\approx1.585
\end{equation}
bits. This is the basis of the widely used ``1.58-bit'' terminology.\cite{bitnet158} However, the information content of an ideal symbol is not the same thing as the storage cost of a complete model. Offsets, scales, residual planes, salience metadata, non-quantized tensors, packing formats, and execution buffers all contribute to the deployed system.

The second distinction is capability. A pretrained model has already accumulated linguistic structure, factual associations, task heuristics, and representations. Aggressive post-training conversion therefore poses a different question from native low-bit training. Native ternary work asks whether a model can be trained under a constrained numerical regime; PTQ asks how much of an existing model survives conversion.\cite{bitnet,bitnet158,catq,ptqtp,twla}

The third distinction is execution. A smaller representation is not automatically a faster representation. Existing work such as AWQ and SmoothQuant demonstrates that hardware-aware transformations and kernels can turn lower precision into real deployment gains, but those gains depend on the target representation and execution stack.\cite{awq,smoothquant} For a custom ultra-low-bit representation, unpacking, dequantization, rotation, auxiliary scales, and kernel overhead can erase the arithmetic advantage.

This paper therefore treats the conversion as a four-part empirical problem:

\begin{enumerate}
\item \textbf{Capability:} what downstream behavior survives?
\item \textbf{Representation:} what is the effective bit budget after adaptive planes and metadata are considered?
\item \textbf{Storage:} what size reduction is achieved by an actual artifact rather than a fake-quantized FP16 checkpoint?
\item \textbf{Execution:} does the compact representation improve inference on the tested hardware?
\end{enumerate}

This framing is intentionally different from a leaderboard-only comparison. The objective is not to claim a new ternarization algorithm. KOTMS, E2M-ATQ, and GPTQ are prior methods. The contribution is an end-to-end empirical characterization of their application to \Qwen{}, together with representation accounting, capability diagnostics, packing analysis, and deployment measurements.

\begin{productbox}
\textbf{Why this matters for an AI product.}
A compressed model is useful only if the representation can be stored, moved, loaded, and eventually executed at an acceptable cost. The present evidence supports a strong storage/compression result and a credible path toward hardware-aware execution, but it does not establish a universal inference-speed or serving-cost advantage. This distinction is central to the product interpretation of the work.
\end{productbox}

\section{Research Positioning and Related Work}

\subsection{Native ternary training}

BitNet established a Transformer architecture in which low-bit weights are part of the training formulation rather than a post-hoc conversion.\cite{bitnet} BitNet b1.58 subsequently popularized ternary weights $\{-1,0,+1\}$ and argued for the practical relevance of approximately 1.58-bit models.\cite{bitnet158} These works provide the conceptual foundation for low-bit language modeling, but their training-time setting differs fundamentally from the present study.

\subsection{General PTQ}

GPTQ introduced one-shot post-training quantization using approximate second-order information to reduce the reconstruction error of large language models.\cite{gptq} AWQ emphasized activation-aware protection of salient weights and hardware-friendly weight-only inference.\cite{awq} SmoothQuant instead transformed activation outliers into weight scaling to enable W8A8 execution.\cite{smoothquant} OmniQuant explored optimization of quantization parameters in a training-free setting.\cite{omniquant}

These methods establish an important principle: calibration and representation design matter, but so does the execution format.

\subsection{Ternary PTQ}

Recent work has moved directly into the ternary regime. PTQTP represents weights using structured ternary planes and explicitly addresses the trade-off between expressiveness and execution complexity.\cite{ptqtp} CAT-Q proposes a post-training ternary conversion with small calibration requirements.\cite{catq} ScaleQ-1.58 further studies calibration for reasoning models using self-generated reasoning traces.\cite{scaleq} TWLA combines KOTMS, E2M-ATQ, and ILA-AMP to target ternary weights and low-bit activations.\cite{twla}

The present work uses TWLA's published components rather than claiming them as new. Importantly, our configuration is W1.58A16: the activation quantization component is deliberately disabled. The paper therefore characterizes the weight-only path and its practical storage/deployment implications rather than reproducing the full W1.58A4 system.

\begin{table}[H]
\centering
\caption{Positioning of this study relative to representative low-bit directions.}
\begin{tabularx}{\linewidth}{l l l X}
\toprule
Work & Regime & Core idea & Relevance to this study\\
\midrule
BitNet & Native low-bit & Train directly with low-bit weights & Establishes native low-bit viability\\
BitNet b1.58 & Native ternary & Ternary weights throughout training & Defines the 1.58-bit ternary target\\
GPTQ & PTQ & Hessian-aware one-shot compensation & Provides the error-compensation paradigm\\
AWQ & PTQ & Activation-aware salient-weight protection & Demonstrates hardware-aware weight-only PTQ\\
PTQTP & Ternary PTQ & Structured trit planes & Closely related representation design\\
CAT-Q & Ternary PTQ & Calibration-efficient ternarization & Closely related pretrained conversion\\
ScaleQ-1.58 & Ternary PTQ & Reasoning-trace calibration & Relevant to future calibration experiments\\
TWLA & Ternary PTQ & KOTMS + E2M-ATQ + ILA-AMP & Direct algorithmic basis of this experiment\\
This work & Empirical system study & Qwen3-4B W1.58A16 characterization & Capability + bit accounting + storage + deployment\\
\bottomrule
\end{tabularx}
\end{table}

\section{Model and Conversion Target}

\subsection{Base model}

The base checkpoint is \Qwen{}, an instruction-tuned 4B-parameter dense model with 36 transformer layers. The experiments were run from the public Qwen3 family checkpoint.\cite{qwen3}

The conversion targeted 252 linear tensors across all 36 layers. The architecture-level accounting in the project record identifies the MLP gate, up, and down projections together with attention $q$, $k$, $v$, and $o$ projections as ternarized. Embeddings and the untied language-model head remain BF16. KOTMS rotation matrices are additional continuous-valued parameters.

\begin{table}[H]
\centering
\caption{Parameter accounting for the converted Qwen3-4B checkpoint.}
\begin{tabular}{lrrr}
\toprule
Component & Parameters & Share & Converted?\\
\midrule
MLP gate projection & 896.5 M & 20.14\% & Yes\\
MLP up projection & 896.5 M & 20.14\% & Yes\\
MLP down projection & 896.5 M & 20.14\% & Yes\\
Attention $q$ projection & 377.5 M & 8.48\% & Yes\\
Attention $o$ projection & 377.5 M & 8.48\% & Yes\\
Attention $k$ projection & 94.4 M & 2.12\% & Yes\\
Attention $v$ projection & 94.4 M & 2.12\% & Yes\\
Embeddings & 389.0 M & 8.74\% & No, BF16\\
LM head & 389.0 M & 8.74\% & No, BF16\\
KOTMS $L,R$ & 40.1 M & 0.90\% & No\\
RMSNorm & 0.2 M & $\sim$0\% & No\\
\midrule
Ternarized linear weights & 3.633 B & 81.62\% & Yes\\
\bottomrule
\end{tabular}
\end{table}

The A16 designation is important: weights are aggressively quantized, but activations, attention operations, and the KV cache remain in FP16. This is not a fully low-bit execution path.

\subsection{Terminology and representation boundary}

The model is referred to as \textit{W1.58A16} throughout the report. When the shorter term ``ternary'' is used, it carries the qualification given in the footnote above: the representation is mixed and adaptive rather than a single ternary plane for every parameter.

The effective bit rate is therefore calculated from the observed use of one or two ternary planes, rather than by simply assigning every weight $\log_2 3$ bits. The measured average is 1.641 bits/weight for the ternarized weight representation.

\section{Method}

\subsection{End-to-end pipeline}

The experimental pipeline is:
\begin{equation}
\text{FP16 Qwen3-4B}
\rightarrow
\text{KOTMS}
\rightarrow
\text{E2M-ATQ}
\rightarrow
\text{GPTQ compensation}
\rightarrow
\text{W1.58A16}.
\end{equation}

KOTMS was reused from a pre-existing rotated checkpoint after its metadata was checked against the published recipe. ILA-AMP was skipped because the experiment used A16 activations. E2M-ATQ performed the actual weight conversion, followed by GPTQ-style error compensation.

\begin{figure}[H]
\centering
\begin{tikzpicture}[node distance=0.9cm, every node/.style={font=\small}]
\node[draw, rounded corners, fill=onebitlight, minimum width=2.5cm, minimum height=0.8cm] (a) {FP16 Qwen3-4B};
\node[draw, rounded corners, fill=onebitlight, minimum width=2.5cm, minimum height=0.8cm, right=of a] (b) {KOTMS rotation};
\node[draw, rounded corners, fill=onebitlight, minimum width=2.5cm, minimum height=0.8cm, right=of b] (c) {E2M-ATQ};
\node[draw, rounded corners, fill=onebitlight, minimum width=2.5cm, minimum height=0.8cm, right=of c] (d) {GPTQ compensation};
\node[draw, rounded corners, fill=onebityellow, minimum width=2.7cm, minimum height=0.8cm, right=of d] (e) {W1.58A16};
\draw[->, thick, onebitteal] (a) -- (b);
\draw[->, thick, onebitteal] (b) -- (c);
\draw[->, thick, onebitteal] (c) -- (d);
\draw[->, thick, onebitteal] (d) -- (e);
\node[below=0.35cm of c, text width=2.8cm, align=center, font=\scriptsize] {ILA-AMP skipped; A16 activations};
\end{tikzpicture}
\caption{Experimental conversion pipeline. The algorithmic components are prior work; the study contribution is their end-to-end characterization on Qwen3-4B.}
\end{figure}
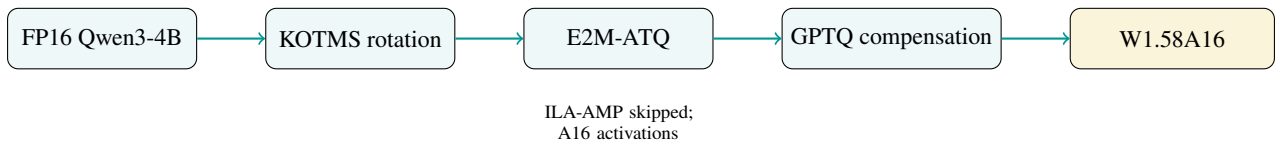

\subsection{KOTMS rotation}

KOTMS uses structured orthogonal matrices to transform the weight space into a ternary-friendly basis. In simplified notation,
\begin{equation}
W' = WR,\qquad R^\top R=I.
\end{equation}
At inference, the equivalent rotation is applied to activations before the quantized linear operation. Because the transformation is orthogonal, it is norm-preserving in exact arithmetic. A prior project measurement reported rotation-only perplexity changing from 8.7998 to 8.8003, consistent with FP16 rounding noise.

For this experiment, a pre-existing rotated checkpoint was reused. Its metadata was verified against the TWLA configuration, including GMM iterations, learning rate, Cayley update, and balance parameters.

\subsection{E2M-ATQ: order-2 asymmetric residual ternarization}

For a masked weight block $W$, the conversion first computes a row-wise offset:
\begin{equation}
\mu = \operatorname{rowmean}(W),\qquad W_c=W-\mu.
\end{equation}
A ternary support is then obtained using
\begin{equation}
\Delta_0=0.75\operatorname{rowmean}(|W_c|)
\end{equation}
and
\begin{equation}
T_{0,i}=
\begin{cases}
+1,&W_{c,i}>\Delta_0,\\
-1,&W_{c,i}<-\Delta_0,\\
0,&\text{otherwise}.
\end{cases}
\end{equation}

The first scale $\alpha_0$ minimizes the squared reconstruction error:
\begin{equation}
\alpha_0=\arg\min_\alpha \|W_c-\alpha T_0\|_2^2.
\end{equation}

The residual is then quantized again:
\begin{equation}
R_1=W_c-\alpha_0T_0,
\end{equation}
\begin{equation}
\Delta_1=0.75\operatorname{rowmean}(|R_1|),
\end{equation}
with a second ternary plane $T_1$ and scale $\alpha_1$.

The reconstructed weight is therefore
\begin{equation}
\widehat W =
\left(\mu+\alpha_0T_0+\alpha_1T_1\right)\odot M,
\end{equation}
where $M$ denotes the applicable salience mask.

Fifteen alternating refinement iterations re-solve $\mu$, $\alpha_0$, and $\alpha_1$ against the current residual while holding the ternary supports fixed. This is the Euclidean-to-manifold refinement step.

The asymmetric offset is not a cosmetic detail. Because $\mu\neq0$, the three values represented by one plane are effectively $\{\mu-\alpha,\mu,\mu+\alpha\}$ rather than a globally centered $\{-\alpha,0,+\alpha\}$ grid. Consequently, exact zeros are not expected to dominate the stored floating reconstruction, even though the underlying discrete supports are ternary.

\subsection{Salience masks and effective bit budget}

The GPTQ wrapper partitions each matrix into four disjoint Hessian-salience-ranked masks under the configuration
\begin{equation}
\texttt{num\_mask}=2(\texttt{num\_p}+1)=4,
\qquad
\texttt{orders}=[2,2,1,1].
\end{equation}

Thus the two most salient masks receive two ternary planes and the remaining masks receive one. On the recorded \texttt{layers.10.mlp.down\_proj.weight} measurement, the mask coverages were 0.5\%, 3.1\%, 10.0\%, and 86.5\%, respectively. The first two masks therefore account for 3.6\% of the measured weights, while 96.5\% use a single plane.

The effective information budget is consequently
\begin{equation}
B_{\mathrm{eff}}
=
(1+f_2)\log_2(3),
\end{equation}
where $f_2$ is the fraction of weights assigned a second plane. The project-wide measured accounting gives
\begin{equation}
B_{\mathrm{eff}}\approx1.641\;\text{bits/weight}.
\end{equation}

This is the number that should be used when discussing the quantized weight representation. It is distinct from the ideal 1.585-bit information content of a single ternary symbol and from the total checkpoint size.

\subsection{GPTQ error compensation}

After ternary approximation, GPTQ-style compensation quantizes weights column-by-column while propagating quantization error into the remaining columns using the inverse Hessian estimated from calibration activations.\cite{gptq}

The experimental configuration used WikiText-2 calibration, 64 samples, sequence length 2048, block size 128, group size 128, \texttt{order2\_group=True}, \texttt{num\_p=1}, and seed 0. The TWLA default damping parameter was \texttt{percdamp}=0.1. A later 0.01 experiment was attempted but did not complete and therefore provides no result.

\section{Experimental Setup}

\subsection{Hardware and software}

All primary measurements were obtained on a single NVIDIA RTX 5070 with 12,227~MiB VRAM under Windows. The recorded software environment was:
\begin{itemize}
\item PyTorch 2.11.0+cu128;
\item CUDA 12.8;
\item Transformers 4.53.2;
\item Datasets 2.16.1;
\item lm-evaluation-harness 0.4.12;
\item NumPy 1.26.4;
\item Python 3.10.11;
\item TWLA commit 805cbb9.
\end{itemize}

The source record also identifies exact artifact hashes for the principal quantized checkpoint and rotated checkpoint. These should be retained in any release package.

\subsection{Calibration}

The model was calibrated on WikiText-2. The principal run used 64 calibration samples with sequence length 2048. The 64-sample choice was constrained by the widest projection's calibration tensor and the practical VRAM ceiling after the execution environment was instrumented.

A critical experimental lesson is that memory reasoning based only on nominal GPU capacity was misleading. A broken \texttt{no\_grad} decorator caused a linear graph-retention leak from 1.99~GiB after the first calibration sample to 10.45~GiB by sample 64. Restoring the decorator reduced the per-layer requirement to below 3~GiB.

\subsection{Perplexity evaluation}

WikiText-2 was evaluated on the full test split using 146 sequences of 2048 tokens. PTB and C4 were subsequently evaluated to test the degree to which WikiText-2 calibration and evaluation might flatter the headline result.

The final source record gives:
\begin{itemize}
\item WikiText-2: 13.639 $\rightarrow$ 18.748;
\item PTB: 24.700 $\rightarrow$ 31.992;
\item C4: 19.831 $\rightarrow$ 28.966.
\end{itemize}

\subsection{Capability evaluation}

Capability was evaluated zero-shot with lm-evaluation-harness. The protocol used a limit of approximately 1500 examples per task, batch size 2, paired per-item records, and 10,000-resample paired bootstrap confidence intervals.

The project explicitly rejected raw accuracy ratios as a retention metric. For fixed-choice tasks, the corrected retention was defined as
\begin{equation}
R=\frac{A_s-B}{A_t-B},
\end{equation}
where $A_s$ is student accuracy, $A_t$ is teacher accuracy, and $B$ is the larger of the chance and majority-class floors.

This correction matters. A model exactly at chance can otherwise appear to retain non-zero ``percentage'' performance simply because the denominator is measured from zero. PIQA provides a concrete example: a raw ratio can be materially higher than the chance-corrected retention.

MMLU was additionally evaluated under both answer-letter and continuation scoring. This is necessary because the two protocols measure different output behaviors and because the earlier Cloe study demonstrated that answer-letter scoring can be distorted by degenerate output policies.

\section{Main Results}

\subsection{Language modelling quality}

\begin{figure}[H]
\centering
\includegraphics[width=0.88\linewidth]{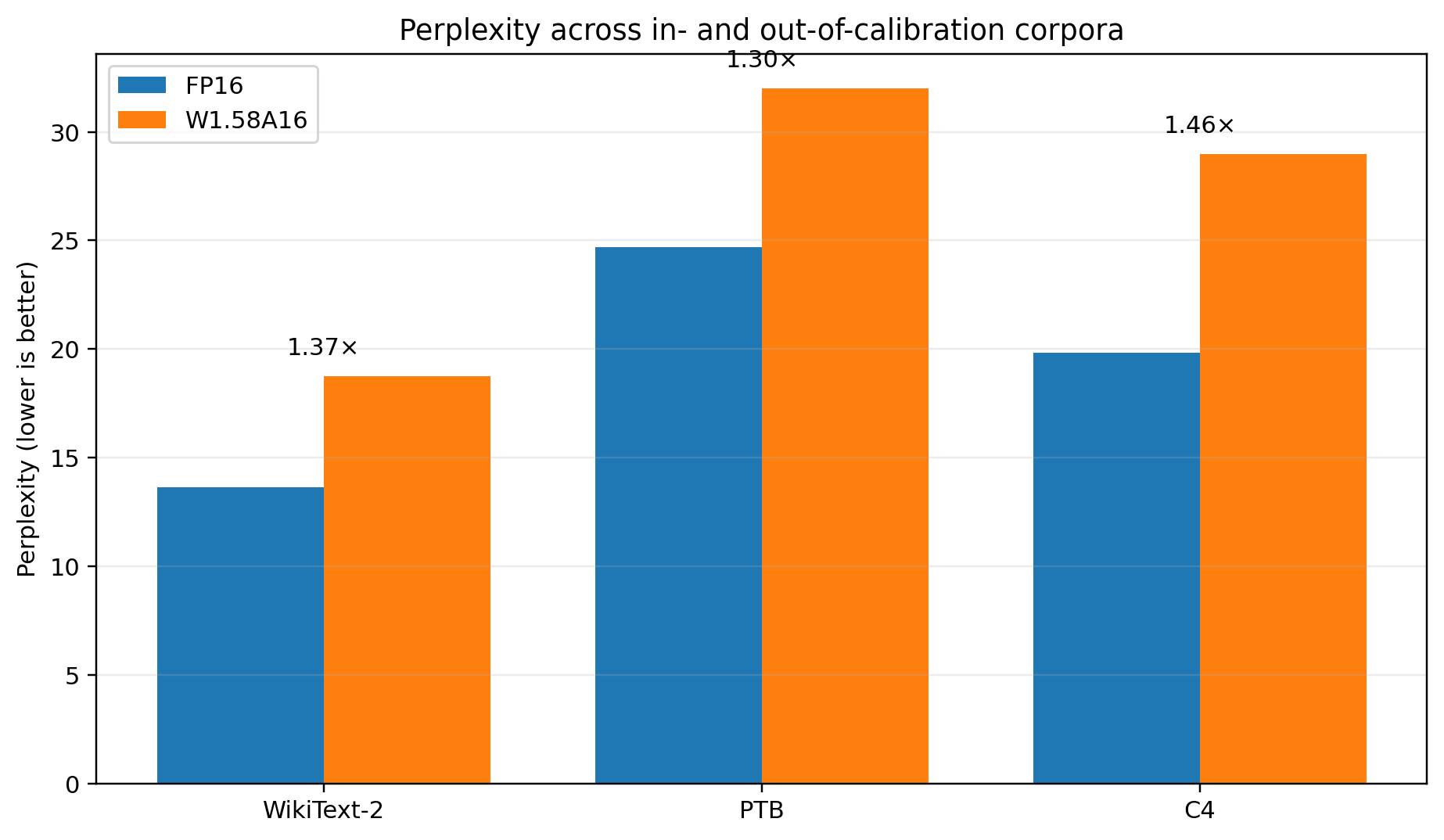}
\caption{Perplexity on three corpora. WikiText-2 is the calibration distribution; PTB and C4 provide out-of-distribution checks.}
\end{figure}

\begin{table}[H]
\centering
\caption{Perplexity across evaluation corpora.}
\begin{tabular}{lrrr}
\toprule
Corpus & FP16 & W1.58A16 & Ratio\\
\midrule
WikiText-2 & 13.639 & 18.748 & 1.375$\times$\\
PTB & 24.700 & 31.992 & 1.295$\times$\\
C4 & 19.831 & 28.966 & 1.461$\times$\\
\bottomrule
\end{tabular}
\end{table}

The important observation is not that one number is ``good'' or ``bad'', but that the degradation depends on the evaluation corpus. WikiText-2 gives the most favorable ratio, while C4 gives the least favorable ratio. This motivates reporting a range rather than presenting 1.375$\times$ as an invariant property of the model.

The C4 result is particularly useful scientifically because it demonstrates a calibration-distribution effect. It does not prove that C4 is universally harder, but it shows that the WikiText-2 headline is not sufficient to characterize language-modelling degradation.

\subsection{Task-level capability}

\begin{figure}[H]
\centering
\includegraphics[width=\linewidth]{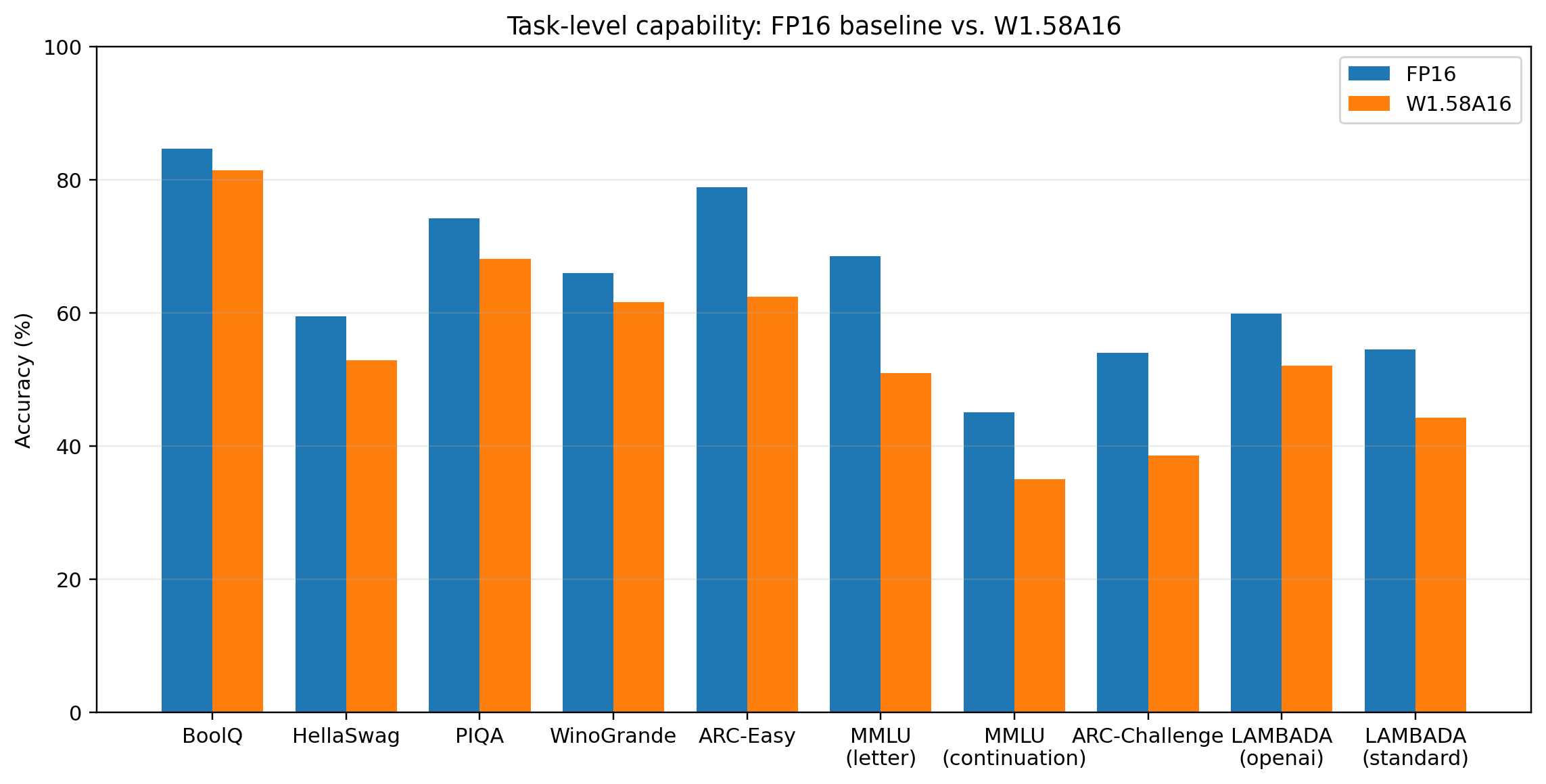}
\caption{Task-level accuracy for FP16 and W1.58A16.}
\end{figure}

\begin{table}[H]
\centering
\scriptsize
\caption{Final task-level comparison. Confidence intervals are paired bootstrap intervals reported by the project.}
\begin{tabular}{lrrrrr}
\toprule
Task & FP16 & W1.58A16 & $\Delta$ pts & 95\% CI & Retention\\
\midrule
BoolQ & 84.7 & 81.4 & -3.3 & [-5.5,-1.3] & 84.6\%\\
HellaSwag & 59.5 & 52.9 & -6.5 & [-8.5,-4.5] & 80.6\%\\
PIQA & 74.2 & 68.1 & -6.1 & [-8.1,-4.1] & 73.9\%\\
WinoGrande & 66.0 & 61.6 & -4.4 & [-7.4,-1.3] & 72.3\%\\
ARC-Easy & 78.9 & 62.4 & -16.5 & [-18.7,-14.3] & 68.5\%\\
MMLU (letter) & 68.5 & 51.0 & -17.5 & point est. & 59.7\%\\
MMLU (continuation) & 45.1 & 35.0 & -10.1 & point est. & 50.0\%\\
ARC-Challenge & 54.0 & 38.6 & -15.4 & [-18.2,-12.5] & 43.8\%\\
LAMBADA-openai & 59.9 & 52.1 & -7.7 & [-9.9,-5.5] & guarded\\
LAMBADA-standard & 54.5 & 44.3 & -10.2 & [-12.5,-7.9] & guarded\\
\midrule
Arithmetic mean & 64.5 & 54.7 & -9.8 & -- & --\\
\bottomrule
\end{tabular}
\end{table}

No reported confidence interval crosses zero. The losses are therefore detectable under the tested paired bootstrap protocol, although the study uses one seed and a finite evaluation limit.

The arithmetic mean of 64.5\% versus 54.7\% is a descriptive average over the ten listed comparisons. It is not a claim of average performance over all language benchmarks, nor a chance-corrected aggregate retention score. The LAMBADA entries are explicitly guarded because an open-vocabulary task does not have the same simple chance floor as fixed-choice multiple-choice tasks.

\subsection{Capability stratification}

\begin{figure}[H]
\centering
\includegraphics[width=\linewidth]{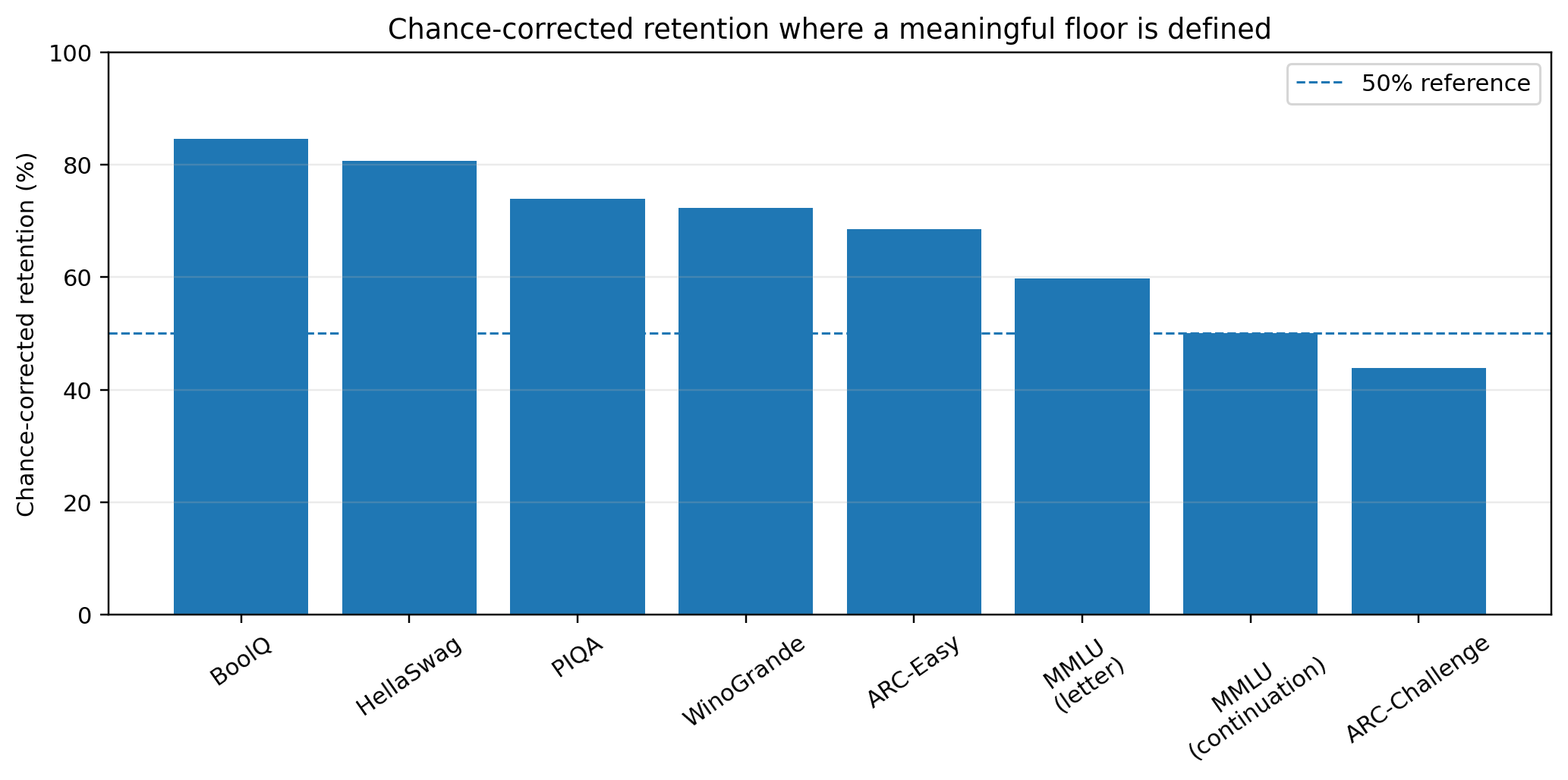}
\caption{Chance-corrected retention for tasks with a meaningful baseline floor. LAMBADA is omitted because the source protocol guards against an invalid floor.}
\end{figure}

The strongest retained behavior is contextual reading and commonsense plausibility. BoolQ retains 84.6\% of teacher performance. HellaSwag, PIQA, and WinoGrande fall in the 72--81\% retention range. ARC-Easy is lower at 68.5\%, while the harder knowledge-oriented tasks fall further: MMLU continuation retains 50.0\%, and ARC-Challenge 43.8\%.

The pattern is therefore stratified rather than uniform. The evidence is consistent with the hypothesis that aggressive post-training discretization disproportionately damages capabilities that depend on precise stored knowledge or fine-grained factual associations. It would be too strong, however, to claim a causal law from this small benchmark suite. The present result is a structured empirical observation requiring confirmation on broader task families.

\begin{figure}[H]
\centering
\includegraphics[width=\linewidth]{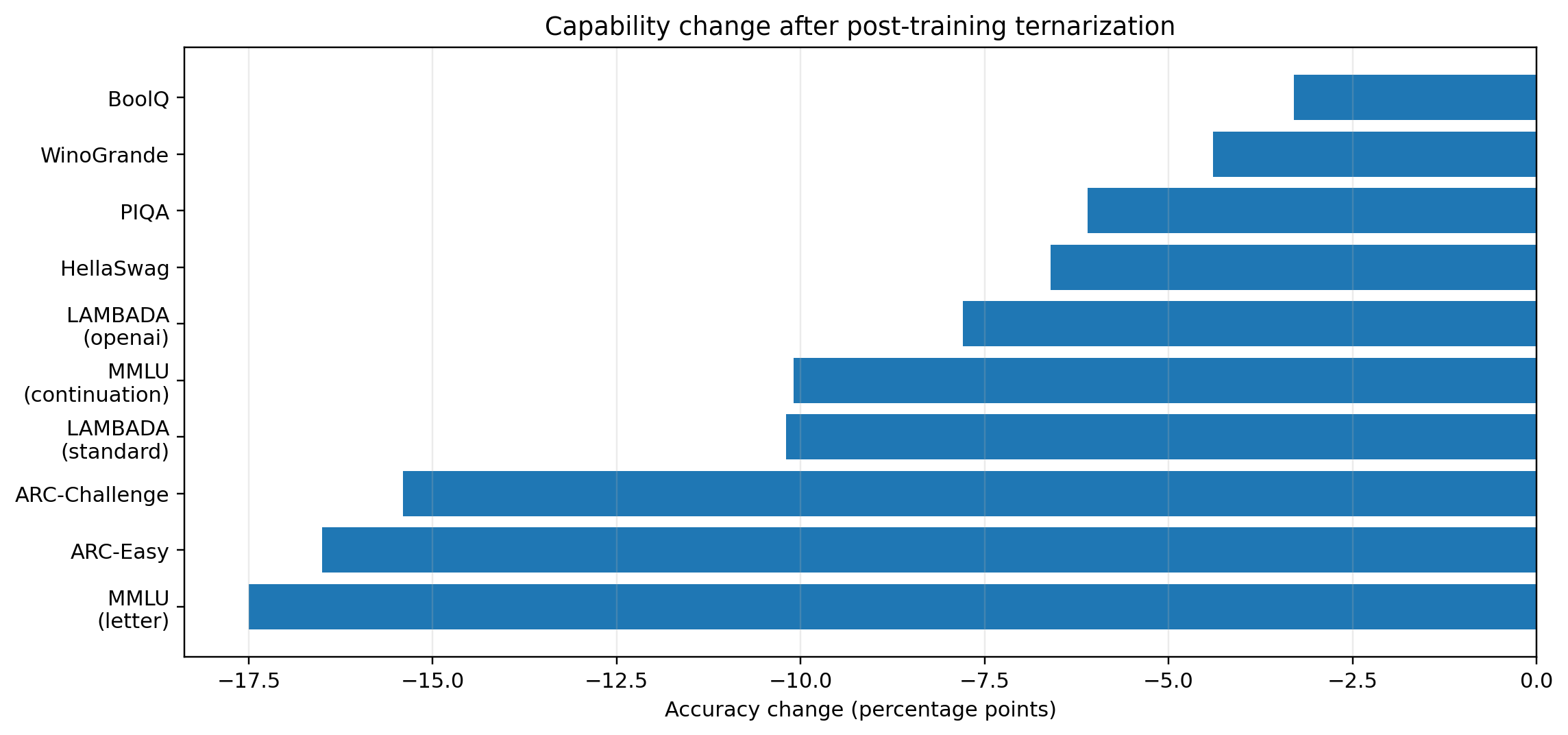}
\caption{Absolute accuracy change by task. The losses are largest on ARC-Easy, MMLU, and ARC-Challenge.}
\end{figure}

\subsection{MMLU scoring protocol}

MMLU is especially sensitive to evaluation protocol. The final record reports:
\begin{itemize}
\item letter scoring: 68.50\% FP16 versus 50.96\% W1.58A16;
\item continuation scoring: 45.10\% FP16 versus 35.04\% W1.58A16.
\end{itemize}

The difference between protocols is substantial for both arms. The absolute FP16--student gap is 23.4 points under letter scoring and 10.1 points under continuation scoring. Consequently, neither MMLU number should be quoted without the scoring protocol.

This is not merely a reporting detail. In extreme quantization, output distributions can become biased toward particular answer formats. A protocol that converts answer formatting into benchmark score can therefore confound capability with decoding behavior.

\section{Representation Analysis}

\subsection{From nominal 1.585 bits to 1.641 effective bits}

The theoretical lower bound of 1.585 bits/weight corresponds to one ternary symbol per weight. The evaluated representation uses a second plane selectively. The measured project-level effective budget is 1.641 bits/weight.

This is an important improvement in accounting precision relative to simply calling the model ``1.58-bit''. The latter is a useful shorthand for the ternary regime, but it does not fully describe the adaptive representation.

\begin{table}[H]
\centering
\caption{Observed salience-mask structure on a recorded projection.}
\begin{tabular}{rrrr}
\toprule
Mask & Planes & Coverage & Bits/weight\\
\midrule
0 & 2 & 0.5\% & 3.17\\
1 & 2 & 3.1\% & 3.17\\
2 & 1 & 10.0\% & 1.58\\
3 & 1 & 86.5\% & 1.58\\
\bottomrule
\end{tabular}
\end{table}

The 3.6\% of weights receiving a second plane are not arbitrary. They are selected by Hessian-based salience. This is conceptually aligned with the broader observation in PTQ that parameter importance is highly non-uniform.\cite{gptq,awq}

\subsection{Why the stored floating tensor can look ``non-ternary''}

A naive inspection of a reconstructed FP16 tensor can report many distinct floating values per row. This does not imply that the discrete representation failed. The representation combines offsets, scales, one or two ternary supports, salience masks, and GPTQ compensation.

In fact, the project observed a particularly useful verification signature: the number of distinct reconstructed values in a recorded row collapsed from 1,169 in the FP16 state to 235 after conversion. The decisive check was end-to-end perplexity: the converted model reproduced approximately 18.748 rather than the FP16 value of 13.639. This guards against accidentally evaluating a rotated or unconverted checkpoint.

\subsection{Representation versus complete checkpoint}

The 1.641 bits/weight figure applies to the targeted weight representation. It does not mean that the complete checkpoint occupies 1.641 bits per parameter.

The complete model retains BF16 embeddings and the language-model head, plus KOTMS rotation matrices and other metadata. The source accounting estimates approximately 745~MiB for packed ternary weights, 1,483~MiB for the untouched embedding/head tensors, and about 78~MiB for the rotation matrices, for a projected total of approximately 2,271~MiB before the later verified packed artifact.

This is why storage measurements must be reported separately from theoretical bit rate.

\newpage
\section{Calibration and Quality Sweep}

\begin{figure}[H]
\centering
\includegraphics[width=0.82\linewidth]{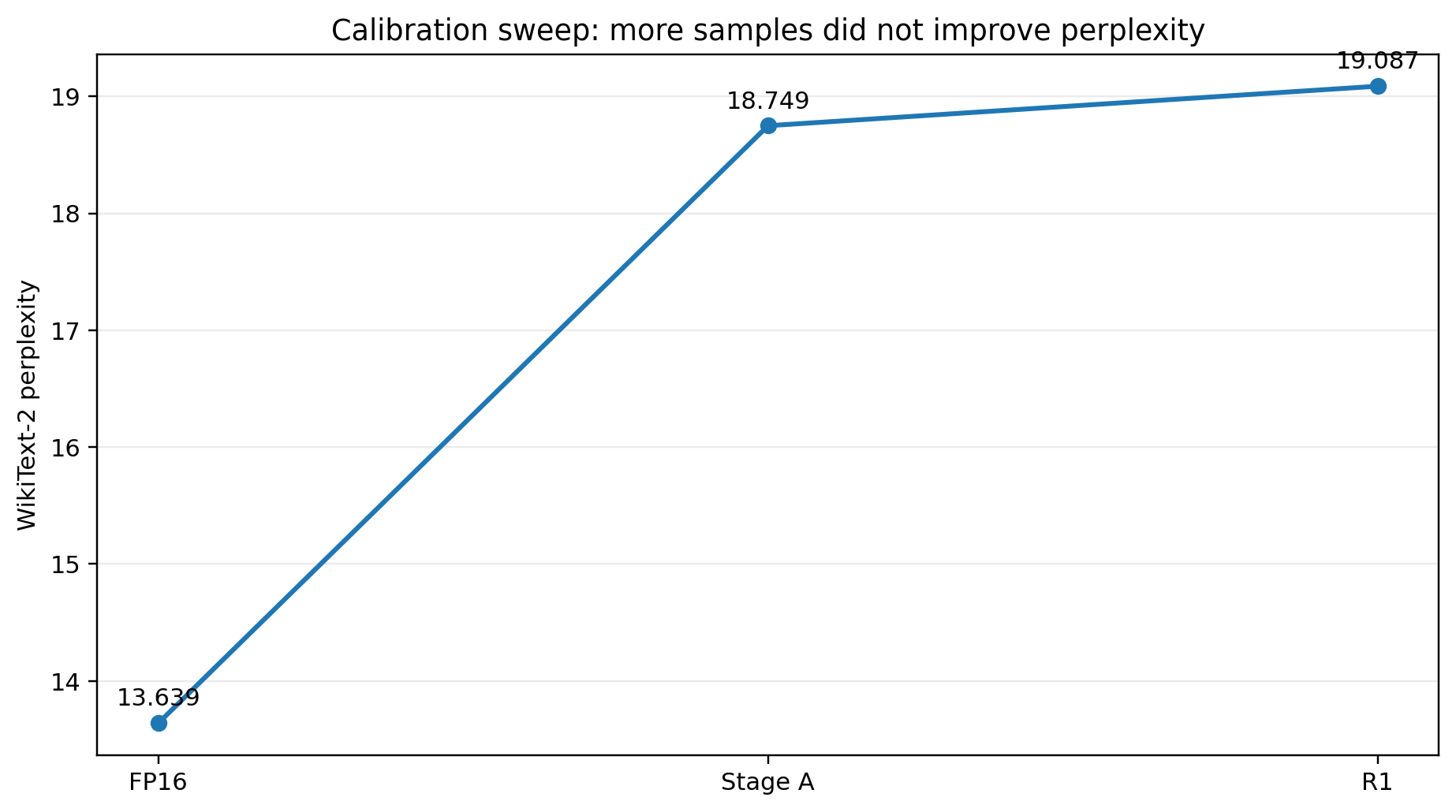}
\caption{Quality sweep on WikiText-2. The R1 increase from 64 to 96 calibration samples worsened perplexity by 1.8\%.}
\end{figure}

The quality sweep changed one variable per run. Stage A used WikiText-2 calibration with 64 samples and \texttt{percdamp}=0.1, producing 18.749 perplexity. R1 increased the calibration sample count to 96 while holding the other configuration fixed, producing 19.087.

This is a negative result, not evidence that 64 samples are inherently better than 96. The source record correctly interprets the 1.8\% difference as ``no improvement from more calibration data'' because the experiment is single-seed and the gap is small.

The remaining levers were not completed in the provided evidence:
\begin{itemize}
\item \texttt{percdamp}=0.01 was exposed but the R2 run paused early;
\item a FineWeb-Edu calibration run was queued but not started;
\item \texttt{num\_p}=2 was not evaluated;
\item \texttt{group\_size}=64 was not evaluated.
\end{itemize}

The report must therefore not imply that the presented configuration is globally optimal. It is the \textbf{best completed configuration in the provided sweep}, not a proven optimum.

\begin{figure}[H]
\centering
\includegraphics[width=0.82\linewidth]{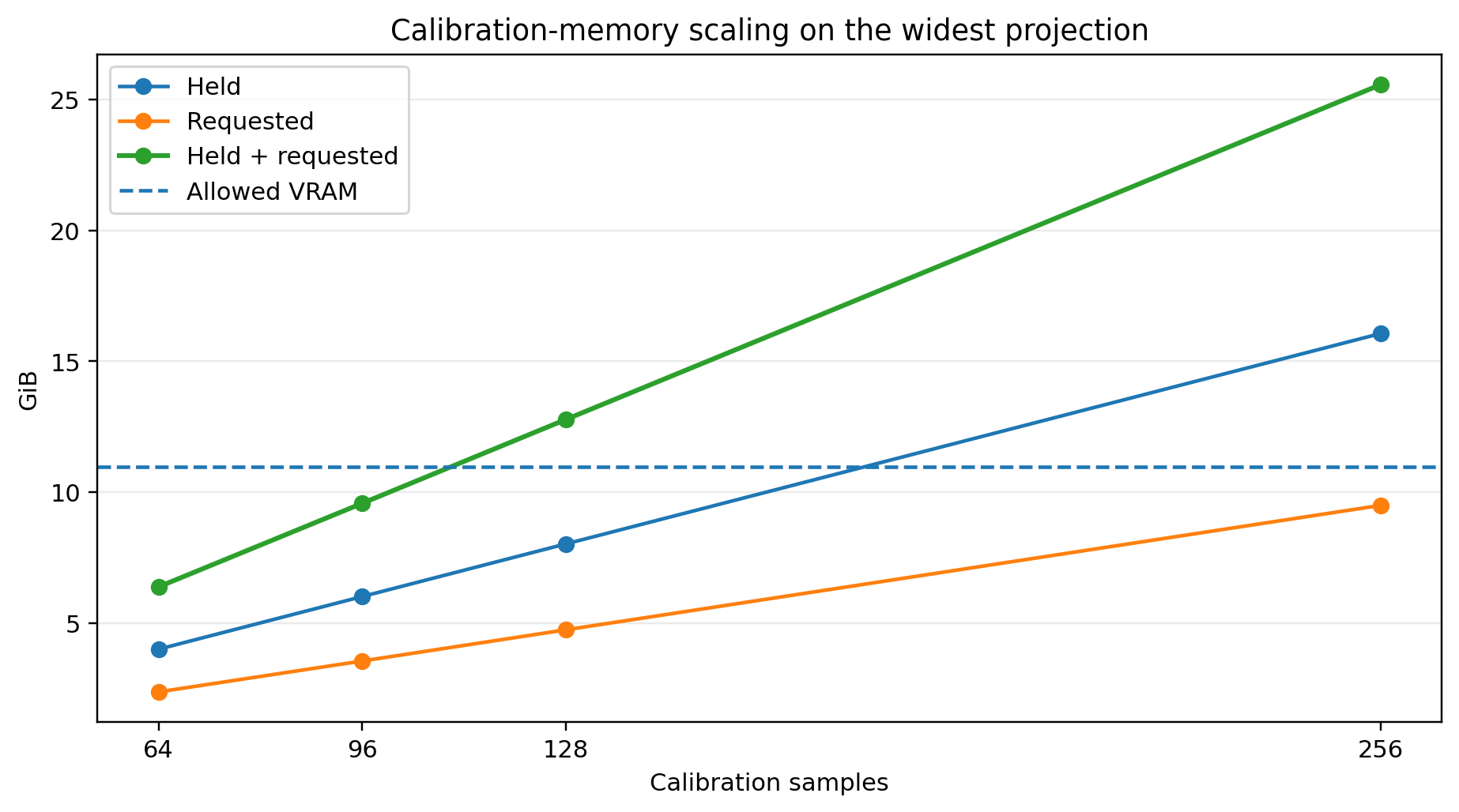}
\caption{Calibration memory scaling for the widest projection. The 128-sample request exceeds the measured allowed VRAM budget.}
\end{figure}

The widest projection requires a calibration tensor whose memory scales linearly with the number of samples. The recorded estimates are 6.39~GiB at 64 samples, 9.58~GiB at 96, 12.78~GiB at 128, and 25.56~GiB at 256. The practical allowed budget was approximately 10.98~GiB. This explains why 128 samples could not be used in the final run under the measured environment.

\section{Storage and Packing}

\subsection{Fake-quantized versus actually packed}

The first saved quantized checkpoint stored the converted values inside FP16 tensors. Consequently, it did not reduce disk footprint despite representing the weights at a much lower effective information rate. This is a critical distinction between \textit{quantization} and \textit{serialization}.

A later packing run persisted the actual ternary planes and their associated scales/offsets using a lattice-aware format. The source record reports:
\begin{table}[H]
\centering
\caption{Recorded packing result.}
\begin{tabular}{lrr}
\toprule
Quantity & Before & Packed\\
\midrule
Checkpoint size & 8.29 GiB & 3.96 GiB\\
WikiText-2 perplexity & 18.344322 & 18.341623\\
PTB perplexity & 34.052238 & 34.042122\\
\bottomrule
\end{tabular}
\end{table}

The packing report also records 0.037\% relative RMSE and a 50-second packing time. Most importantly, the packed artifact reproduced the un-packed model's perplexity to within the reported measurement precision.

\begin{figure}[H]
\centering
\includegraphics[width=0.78\linewidth]{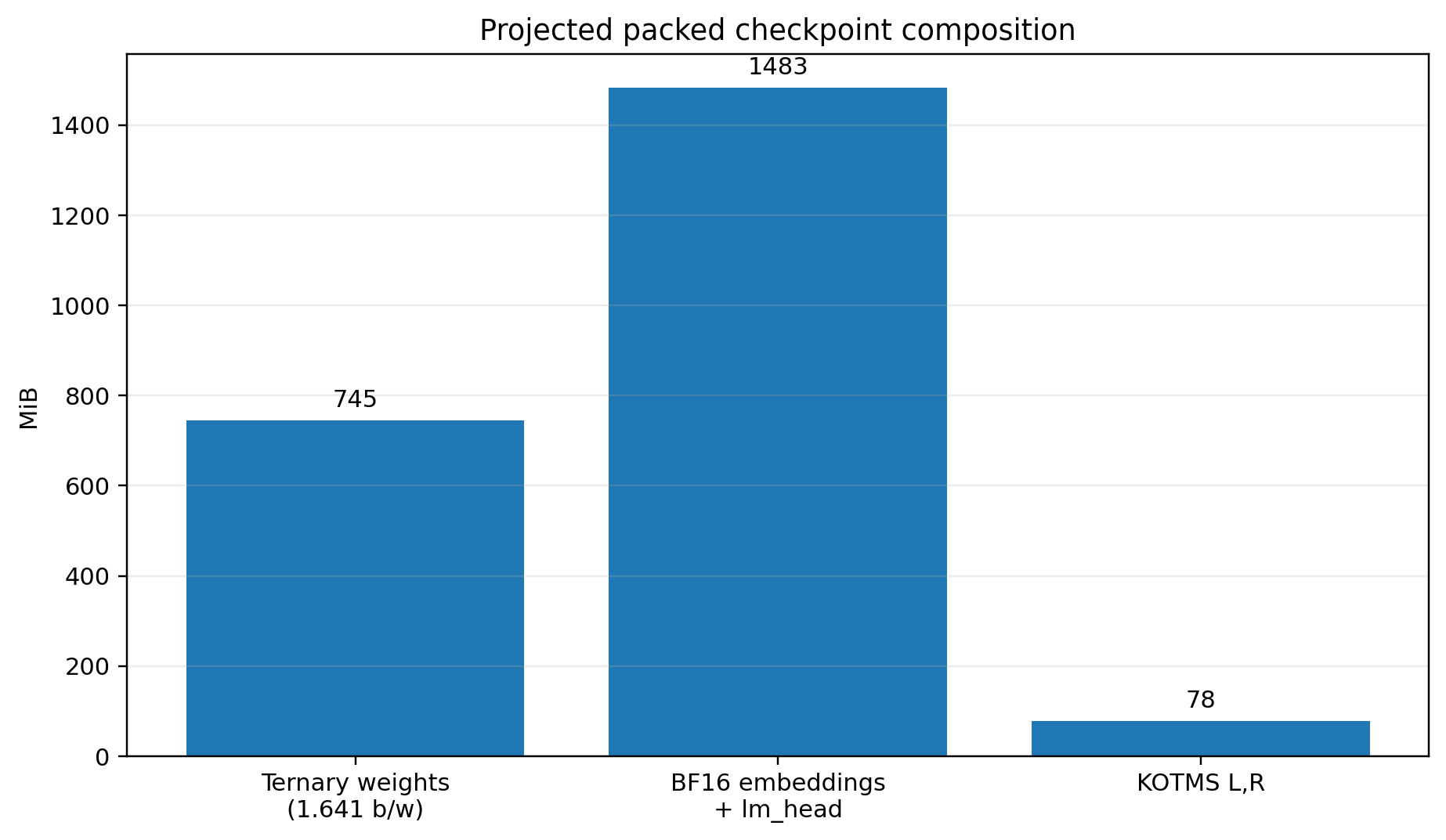}
\caption{Projected composition of the packed checkpoint. The BF16 embedding/head tensors dominate the remaining footprint.}
\end{figure}

The difference between 1.641 bits/weight and 3.96~GiB is therefore not a contradiction. The former describes the adaptive quantized weight representation; the latter describes an entire serialized checkpoint with unquantized components and storage metadata.

\subsection{The excluded lossy packing attempt}

A separate third-party packing artifact used a single 2-bit ternary encoding with FP16 per-group scales. Its recorded whole-checkpoint size was 2.659~GiB, but it introduced a maximum absolute weight error of 0.305 because the second ternary plane and GPTQ residual were discarded.

The artifact had no perplexity or task benchmark. Its own analysis explicitly concluded that it should not be published as-is. We therefore do \textbf{not} use its size or speed as evidence for the primary model claim. This distinction prevents an attractive but unsupported compression number from contaminating the paper.

The underlying engineering lesson is valuable: post-hoc packing cannot reconstruct information that the quantizer discarded at save time. A correct format must persist the discrete supports and their scales during quantization.

\section{Deployment and Execution}

\subsection{Reconstruction-based execution}

The fake-quantized W1.58A16 model was benchmarked directly against FP16. The measured results were:

\begin{table}[H]
\centering
\caption{Measured deployment behavior before native packed execution.}
\begin{tabular}{lrrr}
\toprule
Metric & FP16 & W1.58A16 & Relative\\
\midrule
Load time & 6.57 s & 12.39 s & 1.89$\times$\\
Weight VRAM & 7.545 GiB & 7.620 GiB & 1.01$\times$\\
Prefill & 5,919 tok/s & 5,029 tok/s & 0.85$\times$\\
Decode & 27.41 tok/s & 21.28 tok/s & 0.78$\times$\\
Latency & 36.49 ms/tok & 46.98 ms/tok & 1.29$\times$\\
\bottomrule
\end{tabular}
\end{table}

\begin{figure}[H]
\centering
\includegraphics[width=0.88\linewidth]{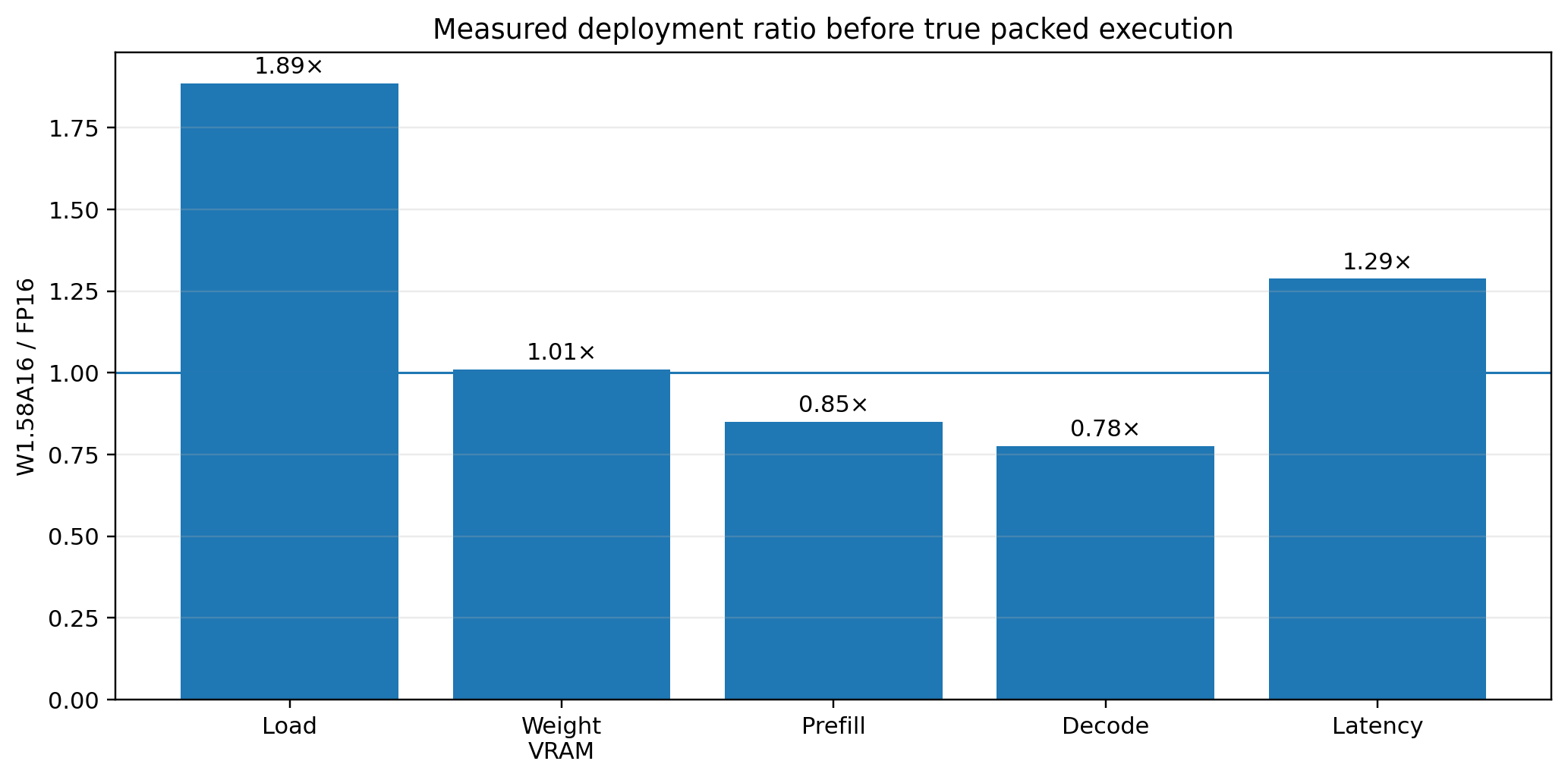}
\caption{Execution ratios relative to FP16. Values above 1 indicate higher cost; values below 1 indicate lower throughput.}
\end{figure}

These results are intentionally not presented as a failure of ternary representations in general. The measured path is a reconstruction-oriented implementation: weights are still stored in FP16, and the KOTMS rotation introduces a real activation-side matrix operation. The experiment therefore measures a specific software stack, not an optimized native ternary accelerator.

This is exactly why storage compression and inference efficiency must remain separate claims.

\subsection{Preliminary packed-kernel microbenchmark}

A separate Triton GEMV kernel was measured on a $4096\times2560$ projection on the RTX 5070:

\begin{table}[H]
\centering
\caption{Preliminary kernel microbenchmark.}
\begin{tabular}{lrr}
\toprule
Path & Time (ms) & Relative to FP16 cuBLAS\\
\midrule
FP16 cuBLAS & 0.0451 & 1.0$\times$\\
Packed Triton & 0.2082 & 4.6$\times$ slower\\
Chunk-unpack + cuBLAS & 0.8924 & 19.8$\times$ slower\\
\bottomrule
\end{tabular}
\end{table}

\begin{figure}[H]
\centering
\includegraphics[width=0.82\linewidth]{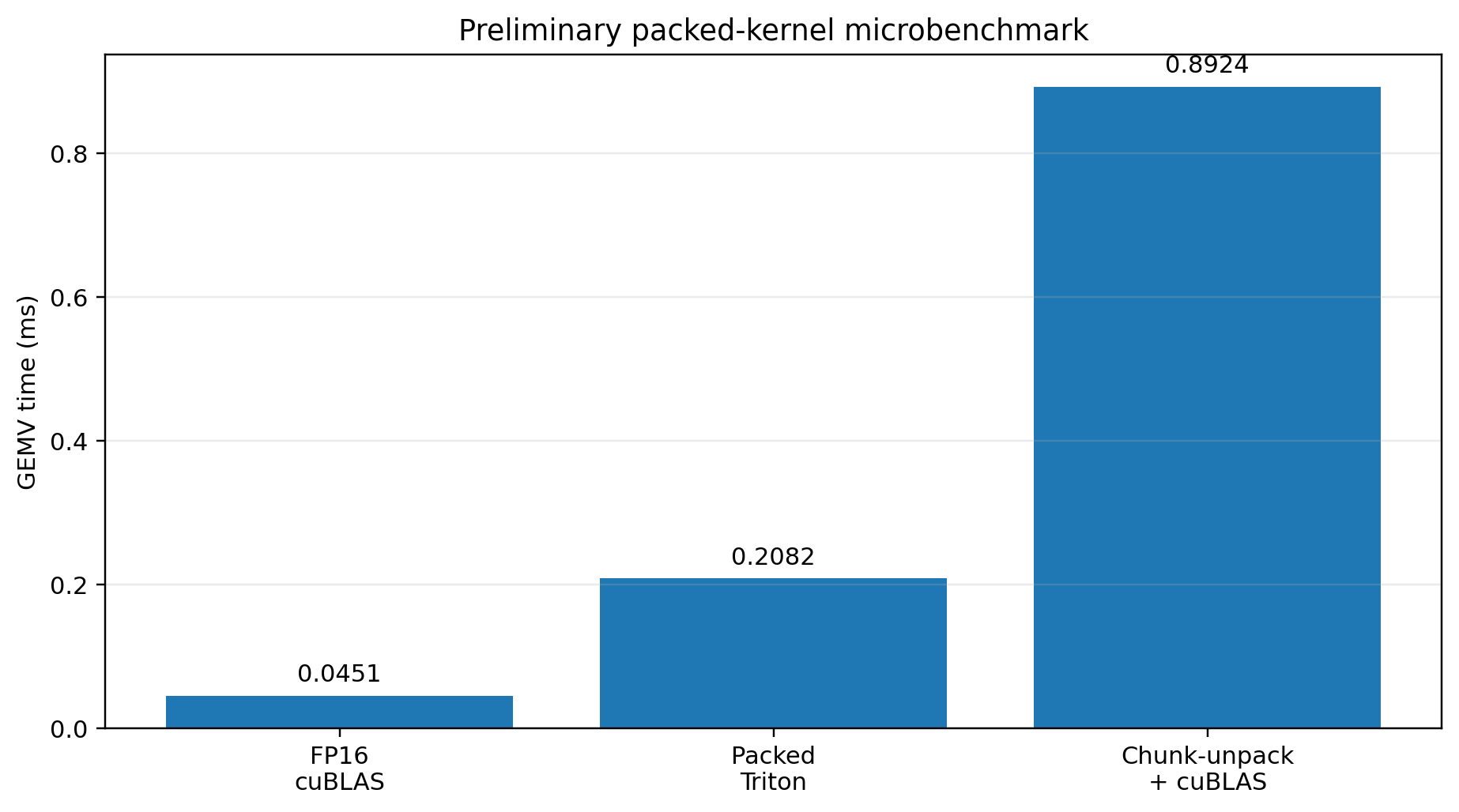}
\caption{Preliminary GEMV timing. The measurement demonstrates a kernel bottleneck, not an intrinsic lower bound on ternary execution.}
\end{figure}

The kernel's own dense-matmul comparison reported approximately 0.00195 maximum absolute discrepancy, indicating that the kernel can reproduce its encoded dense operation to FP16-level numerical precision. The larger 0.305 error reported for the third-party packing artifact belongs to the \textit{encoding} and re-quantization mismatch, not to the Triton kernel's arithmetic.

The systems implication is clear. A compact representation becomes a deployment advantage only when the execution stack consumes that representation directly and efficiently. Existing optimized low-bit systems such as AWQ/TinyChat and BitNet's inference stack demonstrate how much of the practical benefit comes from hardware-aware packing and kernels rather than numerical precision alone.\cite{awq,bitnet158}

\section{Failure Modes as Experimental Boundary Conditions}

The project encountered multiple infrastructure and methodology failures. They are relevant because several could have produced invalid scientific conclusions.

\subsection{Autograd-induced memory leak}

A local checkpointing edit inserted a helper between \texttt{@torch.no\_grad()} and the target function, causing the calibration forward to retain computation graphs. Memory increased from 1.99~GiB after the first sample to 10.45~GiB after 64 samples. The bug was fixed and per-layer memory returned to below 3~GiB.

The scientific lesson is broader than this individual bug: VRAM constraints should be instrumented rather than inferred from model size alone.

\subsection{Windows CUDA spill}

A probe demonstrated that CUDA allocations beyond physical VRAM could spill into system memory under the Windows configuration, producing misleading memory readings and severe performance degradation. The project added a 0.92 per-process memory fraction limit so over-allocation fails cleanly.

\subsection{False ternary artifact}

A pre-existing 8.5~GiB checkpoint appeared to be a candidate ternary artifact but had not actually been ternarized. The verification layer rejected it. Conversely, the real asymmetric E2M-ATQ representation initially looked non-ternary because exact zeros were rare. The project resolved this by inspecting the quantizer's asymmetric offset and enforcing end-to-end perplexity verification.

This is an unusually important reproducibility point. A low-bit experiment should verify both the \textit{representation} and the \textit{forward path}.

\subsection{Power loss and checkpoint recovery}

The machine rebooted approximately 15 minutes before completion of a multi-hour run, losing roughly 2.5 hours of work. Per-layer checkpointing was added so that future interruptions cost approximately one layer rather than the entire run.

\subsection{Benchmark protocol defects}

An adversarial audit caught several issues before final benchmark acceptance:
\begin{enumerate}
\item raw accuracy ratios were rejected as a retention measure because chance floors differ by task;
\item default MMLU letter scoring was identified as potentially misleading;
\item per-item logs were made mandatory;
\item both model arms were evaluated per task to avoid an incomplete paired comparison.
\end{enumerate}

These corrections are part of the experimental method, not merely editorial cleanup.

\newpage
\section{Comparison with the Previous OneBit AI Study}

The previous Cloe study converted Qwen3.5-0.8B using 72.4M tokens of QAT and found a severe loss of specialist knowledge, while still observing measurable downstream learnability and deployment benefits on the tested packed artifact.\cite{onebitcloe}

The present study changes the central conversion strategy from QAT to PTQ. The new model is larger, the conversion requires no gradient training, and the quantizer uses Hessian-aware error compensation. This allows the paper to ask a different question: whether an existing 4B model can be aggressively discretized without spending a large token budget on recovery.

The comparison should not be interpreted as ``PTQ universally beats QAT''. The experiments do not support that claim. The previous Cloe paper itself noted that its QAT budget was limited and that the observed capability gap could not be uniquely separated into irreversible quantization loss and incomplete recovery. The current project also does not provide a controlled QAT/PTQ head-to-head on the same model and compute budget.

What can be said is narrower and stronger: \textbf{the PTQ route produces a working 4B W1.58A16 conversion without full retraining, with substantial retained capability on several tasks and a verified path to multi-gigabyte checkpoint compression.}

The two papers also share an evaluation lesson. Capability should be stratified rather than reduced to a single average. In Cloe, knowledge-heavy behavior approached chance; in the present 4B conversion, knowledge-heavy tasks are degraded but remain substantially above chance on several benchmarks. This suggests that model scale and conversion procedure both matter, but the current evidence is insufficient to isolate their individual causal effects.

\begin{table}[H]
\centering
\caption{High-level comparison with the previous OneBit AI Cloe study.}
\begin{tabularx}{\linewidth}{lXX}
\toprule
Dimension & Previous Cloe study & Present study\\
\midrule
Base model & Qwen3.5-0.8B & Qwen3-4B\\
Conversion & QAT & PTQ\\
Training budget & 72.4M tokens & No training; calibration only\\
Representation & Fully ternary targeted layers & Adaptive order-1/order-2 ternary planes\\
Primary emphasis & Capability stratification + learnability & Capability + bit accounting + storage + execution\\
Main deployment evidence & Packed artifact + throughput & Verified packed artifact; native packed speed remains open\\
Key limitation & Limited recovery budget & Single seed, incomplete tuning/reproduction gate\\
\bottomrule
\end{tabularx}
\end{table}

\section{Scientific Interpretation}

\subsection{Hypothesis assessment}

The motivating hypothesis can be stated as follows:

\begin{quote}
\textit{An aggressively post-training quantized 4B pretrained language model can retain useful downstream capability while reducing its effective weight precision toward the ternary information limit, but practical deployment efficiency will depend on how the representation is serialized and executed.}
\end{quote}

The evidence supports this hypothesis in part.

First, the capability component is supported: the model remains substantially above chance on several tasks, with BoolQ at 81.4\%, PIQA at 68.1\%, and WinoGrande at 61.6\%. At the same time, the hypothesis does not imply negligible degradation; ARC-Challenge and MMLU show substantial losses.

Second, the representation component is supported: the measured quantized weight budget is 1.641 bits/weight, close to the ternary information floor, while 96.5\% of weights in the recorded salience example use a single plane.

Third, the storage component is supported by the later lossless packing record: 8.29~GiB becomes 3.96~GiB with essentially unchanged perplexity.

Fourth, the execution component is \textbf{not yet supported as an advantage}. The reconstruction path is slower than FP16, and the preliminary packed GEMV kernel is slower than cuBLAS. This is not a contradiction of the hypothesis; it identifies the remaining engineering condition for turning representation efficiency into runtime efficiency.

\subsection{What is genuinely new in this study}

The paper does not claim novelty for:
\begin{itemize}
\item ternary weights;
\item the 1.58-bit information limit;
\item KOTMS;
\item E2M-ATQ;
\item GPTQ;
\item generic post-training quantization;
\item native ternary inference as a concept.
\end{itemize}

The defensible contributions are empirical and system-oriented:
\begin{enumerate}
\item an end-to-end Qwen3-4B application of the TWLA weight-only pipeline;
\item explicit effective-bit accounting for adaptive second-plane usage;
\item task-level characterization of capability degradation under aggressive PTQ;
\item cross-corpus perplexity analysis showing calibration-distribution sensitivity;
\item a verified lossless packing path with measured whole-checkpoint reduction;
\item a deployment study showing why the current execution path does not yet convert compression into speed;
\item engineering and evaluation safeguards that prevent several classes of false low-bit results.
\end{enumerate}

This is a meaningful technical characterization even though the underlying quantizer is prior art.

\section{Product and Deployment Implications}

\begin{productbox}
\textbf{The strongest product narrative is compression-first, acceleration-next.}

The current evidence supports a materially smaller model artifact and establishes that a 4B pretrained model can be pushed into a near-ternary weight regime without full retraining. That can matter for model distribution, checkpoint transfer, local storage, and future hardware-aware serving.

The evidence does \emph{not} yet support the stronger statement that the current artifact is cheaper or faster to serve on arbitrary hardware. The measured reconstruction path is slower than FP16, and the preliminary Triton kernel is substantially slower on the tested GEMV. The commercial opportunity therefore lies in combining the demonstrated representation with a native execution stack rather than treating compression itself as the finished product.
\end{productbox}

The most immediate deployment opportunities are:
\begin{enumerate}
\item \textbf{Model distribution:} a 3.96~GiB artifact is materially easier to move and store than an 8.29~GiB checkpoint.
\item \textbf{Memory-constrained deployment:} further quantization of the BF16 embedding/head tensors could reduce the remaining footprint.
\item \textbf{Native kernels:} direct consumption of the stored ternary planes can remove reconstruction overhead.
\item \textbf{Hardware co-design:} the low-cardinality arithmetic structure is compatible with specialized low-bit kernels and accelerators, as demonstrated conceptually by prior BitNet systems.\cite{bitnet158}
\item \textbf{Specialized models:} the capability profile suggests that broad general-purpose replacement is not yet justified, while targeted workloads may offer a more favorable quality-efficiency trade-off.
\end{enumerate}

\subsection{What the current evidence does not establish}

The following claims should not be made from this dataset:
\begin{itemize}
\item universal inference speedup;
\item universal serving-cost reduction;
\item equivalence to a native 1.58-bit model;
\item state-of-the-art capability among ternary models;
\item superiority over every alternative PTQ method;
\item production robustness across hardware;
\item general scaling laws across model sizes.
\end{itemize}

The study is strongest when it treats these as future engineering and research questions rather than as conclusions already settled.

\section{Limitations and Required Next Experiments}

The principal limitations are:

\begin{enumerate}
\item \textbf{No independent reproduction gate on Qwen3-4B.} The source record notes that Qwen3-4B is absent from the TWLA paper. A reproduction gate on the authors' published reference model remains the highest-value validation.
\item \textbf{Single seed.} Capability results do not quantify run-to-run variation.
\item \textbf{Incomplete tuning.} The \texttt{percdamp}=0.01 run did not complete; alternative calibration corpora and representation parameters were not evaluated.
\item \textbf{Incomplete packed-model capability evaluation.} The later packing record demonstrates perplexity preservation but the provided source set does not contain an end-to-end packed-model task benchmark.
\item \textbf{Preliminary kernel measurement.} The Triton GEMV is a prototype and should not be treated as a lower bound on native ternary performance.
\item \textbf{Partial parameter quantization.} Embeddings and the LM head remain BF16 and account for a substantial fraction of the remaining artifact.
\item \textbf{Limited benchmark breadth.} The final task suite is informative but smaller than comprehensive LLM evaluations.
\end{enumerate}

The highest-value next experiments are therefore:
\begin{enumerate}
\item complete the published-method reproduction gate;
\item complete \texttt{percdamp}=0.01;
\item rerun with a broader calibration corpus such as FineWeb-Edu;
\item benchmark the exact packed artifact end-to-end;
\item persist the discrete lattice directly during quantization and build a lossless native kernel;
\item evaluate embedding/head quantization;
\item repeat the capability suite across seeds;
\item evaluate the same conversion procedure on at least one additional model scale.
\end{enumerate}

\section{Conclusion}

This study demonstrates that a pretrained Qwen3-4B model can be pushed into an aggressive post-training low-bit regime with a measured 1.641-bit/weight quantized representation and substantial retained capability on a subset of downstream tasks. The result is not uniform preservation: contextual and commonsense tasks are more robust than knowledge-heavy tasks, and the mean accuracy cost across the ten reported comparisons is 9.8 percentage points.

The storage story is stronger than the runtime story. A later lattice-aware packing run reduces the checkpoint from 8.29~GiB to 3.96~GiB while preserving the recorded perplexity to measurement precision. This establishes that the representation can become a materially smaller artifact rather than merely a fake-quantized FP16 checkpoint.

Execution remains the key unresolved systems problem. The reconstruction-based path is slower than FP16, and the preliminary packed Triton GEMV is 4.6$\times$ slower than FP16 cuBLAS on the tested shape. These measurements should not be generalized into a claim that ternary arithmetic is intrinsically slower; they show that \textit{the current software path is not yet hardware-efficient}.

The strongest interpretation of the work is therefore neither ``ternarization solves efficient inference'' nor ``ternarization destroys model capability.'' The evidence supports a more useful engineering conclusion: \textbf{near-ternary post-training conversion can create a substantially smaller 4B model artifact while retaining meaningful portions of pretrained capability, but turning that representation into a production efficiency advantage requires native packing, kernels, and broader validation.}

\section*{Reproducibility Note}

All quantitative claims in this report are derived from the provided OneBit AI project records. The primary completed conversion used Qwen3-4B, KOTMS rotation, W1.58A16, 64 calibration samples, sequence length 2048, WikiText-2 calibration, \texttt{percdamp}=0.1, block size 128, group size 128, \texttt{order2\_group=True}, \texttt{num\_p=1}, and seed 0. The principal environment was Windows with an NVIDIA RTX 5070 (12,227~MiB), PyTorch 2.11.0+cu128, Transformers 4.53.2, Datasets 2.16.1, lm-eval 0.4.12, and Python 3.10.11.

The report deliberately excludes unsupported results from the third-party lossy packed artifact. It also treats the unfinished \texttt{percdamp}=0.01 experiment as unfinished rather than assigning it a result.

\appendix

\section{Complete Experimental Ledger}
\vspace{-0.3cm}
\begin{longtable}{p{0.17\linewidth}p{0.22\linewidth}p{0.18\linewidth}p{0.31\linewidth}}
\toprule
Stage & Configuration & Outcome & Interpretation\\
\midrule
Stage A & WikiText-2, 64 samples, pd=0.1 & 18.749 ppl & Best completed configuration\\
R1 & WikiText-2, 96 samples, pd=0.1 & 19.087 ppl & Negative result; no improvement from more samples\\
R2 & WikiText-2, 64 samples, pd=0.01 & Incomplete & No result; must not be cited as an experiment outcome\\
R3 & Alternative calibration & Not started & Open experiment\\
Packing A & Lattice-aware serialization & 8.29 GiB $\rightarrow$ 3.96 GiB; PPL preserved & Primary packing result\\
Packing B & Third-party 2-bit re-quantization & 2.659 GiB; max error 0.305 & Excluded from primary artifact claim\\
Kernel & Packed Triton GEMV & 0.2082 ms & 4.6$\times$ slower than FP16 cuBLAS on one shape\\
\bottomrule
\end{longtable}

\section{Capability Table with Baselines}
\vspace{-0.3cm}
\begin{longtable}{lrrrrrr}
\toprule
Task & Baseline & FP16 & W1.58A16 & $\Delta$ & CI & Retention\\
\midrule
BoolQ & 63.1 & 84.7 & 81.4 & -3.3 & [-5.5,-1.3] & 84.6\\
HellaSwag & 25.8 & 59.5 & 52.9 & -6.5 & [-8.5,-4.5] & 80.6\\
PIQA & 50.7 & 74.2 & 68.1 & -6.1 & [-8.1,-4.1] & 73.9\\
WinoGrande & 50.0 & 66.0 & 61.6 & -4.4 & [-7.4,-1.3] & 72.3\\
ARC-Easy & 26.4 & 78.9 & 62.4 & -16.5 & [-18.7,-14.3] & 68.5\\
MMLU (letter) & 25.0 & 68.5 & 51.0 & -17.5 & point & 59.7\\
MMLU (cont.) & 25.0 & 45.1 & 35.0 & -10.1 & point & 50.0\\
ARC-Challenge & 26.5 & 54.0 & 38.6 & -15.4 & [-18.2,-12.5] & 43.8\\
LAMBADA-openai & -- & 59.9 & 52.1 & -7.7 & [-9.9,-5.5] & guarded\\
LAMBADA-standard & -- & 54.5 & 44.3 & -10.2 & [-12.5,-7.9] & guarded\\
\bottomrule
\end{longtable}

\section{Implementation and Reproducibility Checklist}

\begin{itemize}
\item Base model revision: Qwen3-4B, instruction-tuned.
\item Quantization method: TWLA E2M-ATQ with KOTMS rotation and GPTQ compensation.
\item Activations: FP16/A16.
\item Calibration: WikiText-2, 64 samples, 2048 tokens.
\item Quantization damping: 0.1.
\item Block/group size: 128.
\item Order-2 salience grouping: enabled.
\item Number of additional-plane groups: \texttt{num\_p=1}.
\item Seed: 0.
\item Evaluation: full WikiText-2 test split for PPL; zero-shot benchmark suite for capability.
\item Statistical method: paired bootstrap, 10,000 resamples.
\item Hardware: NVIDIA RTX 5070, 12,227~MiB, Windows.
\item Software: PyTorch 2.11.0+cu128; Transformers 4.53.2; Datasets 2.16.1; lm-eval 0.4.12; NumPy 1.26.4; Python 3.10.11.
\item Quantized checkpoint hash prefix: \texttt{13B6608425DF3245}.
\item Rotated checkpoint hash prefix: \texttt{DE3769B658762A2F}.
\end{itemize}

\section{Interpretation Guardrails}

The following statements are recommended for any future summary of this work:

\begin{enumerate}
\item ``The model uses an adaptive ternary weight representation with a measured effective budget of 1.641 bits/weight.''
\item ``The completed conversion retains substantial performance on several commonsense and contextual tasks, while knowledge-heavy tasks degrade more strongly.''
\item ``A later lattice-aware packing run reduces the recorded checkpoint from 8.29~GiB to 3.96~GiB with essentially unchanged perplexity.''
\item ``The current execution path is slower than FP16; native hardware-efficient execution remains future work.''
\end{enumerate}

Avoid:
\begin{enumerate}
\item ``1.58-bit means the whole checkpoint is 1.58 bits/parameter.''
\item ``The packed model is faster'' without a full end-to-end benchmark of the packed artifact.
\item ``The method is state of the art'' based on these experiments.
\item ``64 calibration samples are better than 96'' from the single R1 comparison.
\item ``PTQ is better than QAT'' based only on comparison with the previous Cloe study.
\end{enumerate}

\newpage
\bibliographystyle{unsrtnat}
\bibliography{cloe_v2_references}

\end{document}